\documentclass{article}

\usepackage[preprint]{corl_2026} 

\usepackage{booktabs}
\usepackage{multirow}
\usepackage[table]{xcolor}
\usepackage{graphicx}
\usepackage{caption} 
\usepackage{amsmath}
\usepackage{amsfonts}
\usepackage{wrapfig}
\usepackage{enumitem}

\title{PoLAR: Factorizing Extent and Mode in \\Latent Actions for Robot Policy Learning}

\author{
  Youngjoon Jeong \quad Jihwan Yu \quad Minsoo Jo \quad Junha Chun \quad Taesup Kim\thanks{Corresponding author.}\\
  Seoul National University
}
\begin{document}

\makeatletter
\let\@oldmaketitle\@maketitle
\renewcommand{\@maketitle}{%
  \@oldmaketitle%
  \vspace{-2.5em} 
  \begin{center}
    \includegraphics[width=1.0\textwidth]{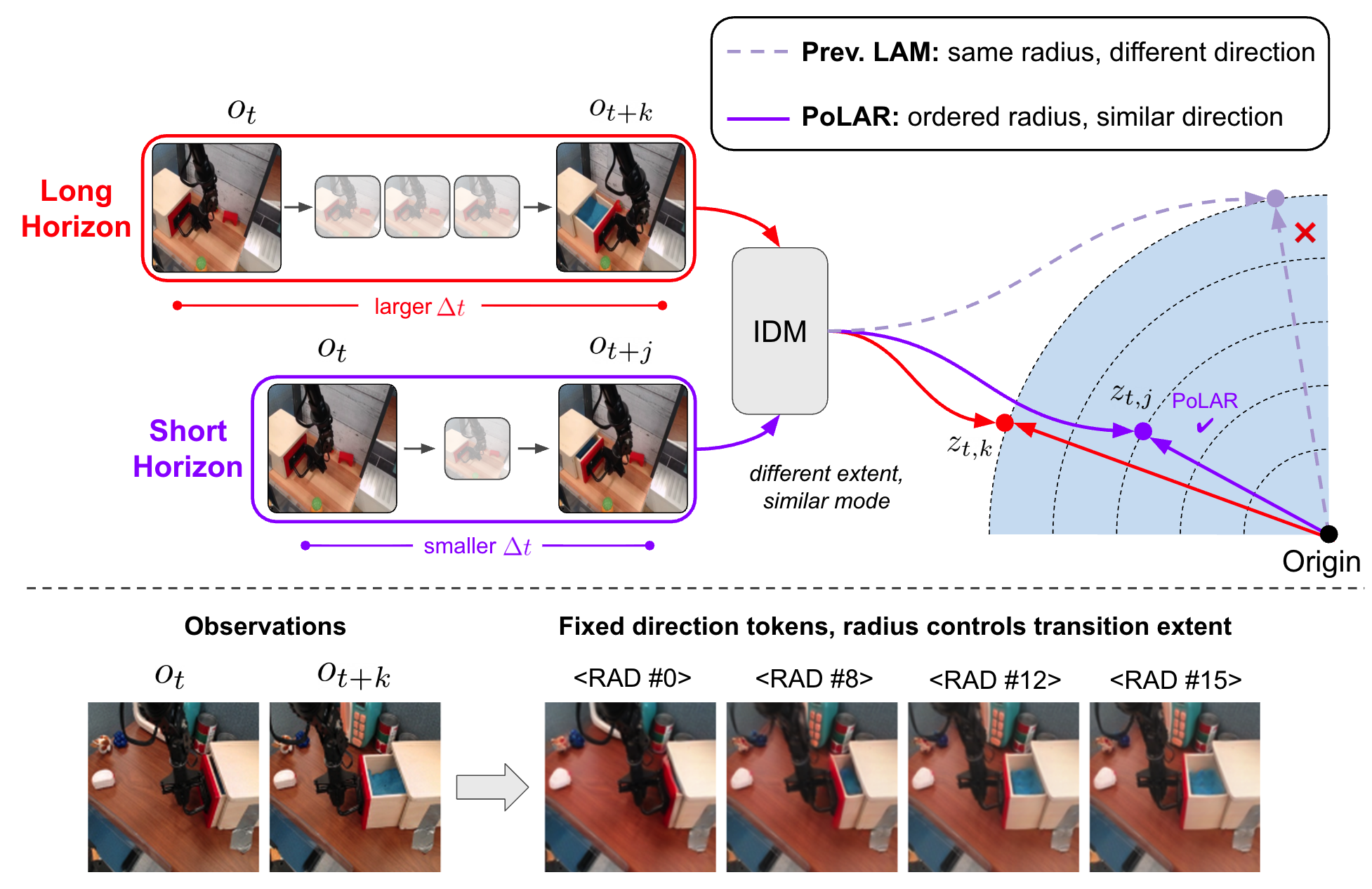}
    \captionof{figure}{\textbf{PoLAR factorizes transition extent and mode in latent actions.}
PoLAR uses temporal offset to order transition extent along radius, allowing similar transition modes to remain in similar directions in latent actions.
Sweeping the radius token with fixed direction increases decoded transition extent.}
    \label{fig:teaser}
  \end{center}
  \vspace{-1.0em} 
}
\makeatother

\maketitle


\begin{abstract}
Robot actions vary in both extent and mode: how much change they produce and what kind of change they produce.
A latent action space that reflects this structure may make it easier for downstream policies to map latent actions to robot actions.
However, existing latent action methods typically rely on visual reconstruction from observation pairs without explicitly encouraging such structure.
We introduce \textbf{Po}lar \textbf{L}atent \textbf{A}ctions with \textbf{R}adial structure (PoLAR), which imposes a radial-direction structure on latent actions, encouraging radius to encode transition extent and direction to retain transition mode.
PoLAR uses temporal offsets between observations as a weak proxy for transition extent, encouraging transitions with larger offsets to occupy larger radii.
We instantiate this structure in hyperbolic space, whose expanding volume with radius offers a natural fit for more diverse transition modes at larger extents.
Across in-task and large-scale pretraining settings, PoLAR improves downstream policy performance in simulation and real-world robot experiments, outperforming latent action baselines and strong pretrained VLAs.
These results suggest that structuring latent actions by extent and mode helps transfer visual pretraining to robot policy learning.
\end{abstract}



\section{Introduction}

Latent actions summarize the change between two observations as a compact representation~\citep{lapo,ye2024latent,bu2025univla,bruce2024geniegenerativeinteractiveenvironments,garrido2026learninglatentactionworld}.
An inverse dynamics model observes two frames from the same trajectory and compresses the transition between them into a bottlenecked representation.
Because this representation is inferred from an observation pair rather than either frame alone, it is encouraged to capture information about the transition rather than static appearance.
The resulting representation describes visual change and helps downstream policies connect observations to low-level robot actions using action-labeled trajectories~\citep{ye2024latent,bu2025univla,chen2025villa0x0,jeong2026learningactrobustlyviewinvariant,lee2026mvplam}.

Robot actions vary in both extent and mode: how much change they produce and what kind of change they produce.
A latent action space that reflects this structure may make it easier for downstream policies to predict latent actions and map them to robot actions.
However, prior latent action methods typically encode each visual transition into a continuous latent vector~\citep{jeong2026learningactrobustlyviewinvariant,liang2025clamcontinuouslatentaction,nikulin2025latent} or a discrete token sequence~\citep{ye2024latent,bu2025univla,chen2025villa0x0,lee2026mvplam}, without explicitly organizing the representation by extent and mode.
In particular, short and long transitions of same action mode are not explicitly encouraged to remain related in the latent space.
This motivates a shared structure in which related transitions can differ in extent while retaining a similar mode.

In this paper, we introduce \textbf{Po}lar \textbf{L}atent \textbf{A}ctions with \textbf{R}adial structure (\textbf{PoLAR}), a latent action learning framework that equips latent actions with a polar geometry.
Rather than encoding visual change as an undifferentiated latent code, PoLAR uses the radius to represent transition extent and the direction to distinguish transition mode.
PoLAR uses the temporal offset between observations as an ordinal proxy for transition extent, encouraging transitions with larger offsets to occupy larger radii.
By comparing short and long transitions from the same starting observation, PoLAR encourages latent actions to focus on what changes between observations.
This radial bias reduces the pressure for the direction to absorb extent-related information, encouraging it to retain information about transition mode.
We instantiate this structure in hyperbolic space, where angular capacity grows with radius~\citep{nickel2017poincareembeddingslearninghierarchical,ganea2018hyperbolicneuralnetworks,Ge_2023_CVPR,pmlr-v202-desai23a}, providing additional capacity for transition modes at larger extents.
We evaluate PoLAR across continuous and discrete latent action parameterizations, in-task and large-scale pretraining regimes, and simulated and real-world manipulation tasks.
PoLAR consistently improves downstream policy learning over conventional latent action baselines and strong pretrained VLAs.
Figure~\ref{fig:teaser} provides an overview of the framework.

\section{Related Work}

\noindent\textbf{Latent actions.}
Latent action models encode observation-to-observation transitions as bottlenecked continuous latents~\citep{liang2025clamcontinuouslatentaction,nikulin2025latent,jeong2026learningactrobustlyviewinvariant} or vector-quantized discrete tokens~\citep{lapo,ye2024latent,bu2025univla,bruce2024geniegenerativeinteractiveenvironments,chen2025villa0x0,jang2025dreamgenunlockinggeneralizationrobot,nvidia2025gr00tn1openfoundation,lee2026mvplam,oord2018neuraldiscreterepresentationlearning}.
These representations support world models~\citep{bruce2024geniegenerativeinteractiveenvironments,garrido2026learninglatentactionworld,gao2025adaworldlearningadaptableworld,jiang2026olaf} and policies learned from action-free videos or cross-embodiment data~\citep{ye2024latent,bu2025univla,chen2025villa0x0,lee2026mvplam, bauer2025latentactiondiffusioncrossembodiment,agibotworldcontributors2025agibotworldcolosseolargescale}.
Most prior methods learn a single transition code in which transition extent and transition mode can be entangled.
PoLAR instead structures the latent action space so that transition extent is represented radially, encouraging angular directions to distinguish transition modes.

\noindent\textbf{Temporal structure as weak supervision.}
Temporal order is a useful source of weak supervision in sequential observation data, requiring neither low-level action labels nor simulator states.
Prior work uses temporal structure for frame-level alignment and phase representations~\citep{sermanet2018timecontrastivenetworksselfsupervisedlearning,dwibedi2019temporalcycleconsistencylearning}, robot-oriented visual or reward pretraining~\citep{nair2022r3muniversalvisualrepresentation,ma2023livlanguageimagerepresentationsrewards,ma2023vipuniversalvisualreward}, progress modeling from passive videos~\citep{yang2024rank2rewardlearningshapedreward}, and representation learning for offline policy pretraining~\citep{park2024foundationpolicieshilbertrepresentations}.
PoLAR instead uses temporal order to structure the geometry of transition-level latent actions: in temporally coherent manipulation trajectories, larger temporal gaps often correspond to larger robot, object, or task-state changes, providing a weak ordinal proxy for transition extent.

\noindent\textbf{Representation geometry.}
Representation geometry can assign different roles to direction and norm: angular separation often carries discriminative semantics~\citep{Wang_2017,liu2018spherefacedeephypersphereembedding,wang2018cosfacelargemargincosine,Deng_2022,wang2022understandingcontrastiverepresentationlearning}, while feature norm can encode non-semantic quantities such as familiarity, reliability, or information content~\citep{park2023understandingfeaturenormoutofdistribution,liu2019largescalelongtailedrecognitionopen,oyama2023normwordembeddingencodes}.
Hyperbolic geometry provides radial capacity: volume grows exponentially with radius, so larger-radius shells support more angular distinctions than Euclidean space.
This property has supported structured representation learning~\citep{nickel2017poincareembeddingslearninghierarchical,ganea2018hyperbolicentailmentconeslearning,ganea2018hyperbolicneuralnetworks,Ge_2023_CVPR,pmlr-v202-desai23a} and recent decision-making, world-modeling, and robustness settings~\citep{cetin2022hyperbolicdeepreinforcementlearning,klein2026understandingimprovinghyperbolicdeep,zhang2026geoworldgeometricworldmodels,Jo_2026}.
For latent actions, this motivates a radial geometry: longer-horizon transitions can involve more diverse object motions, contacts, and task-state changes.
PoLAR therefore encourages transition extent to be represented by radius, leaving direction to distinguish transition modes.
\section{Methods}

\begin{figure*}[t]
\centering
\includegraphics[width=0.85\textwidth]{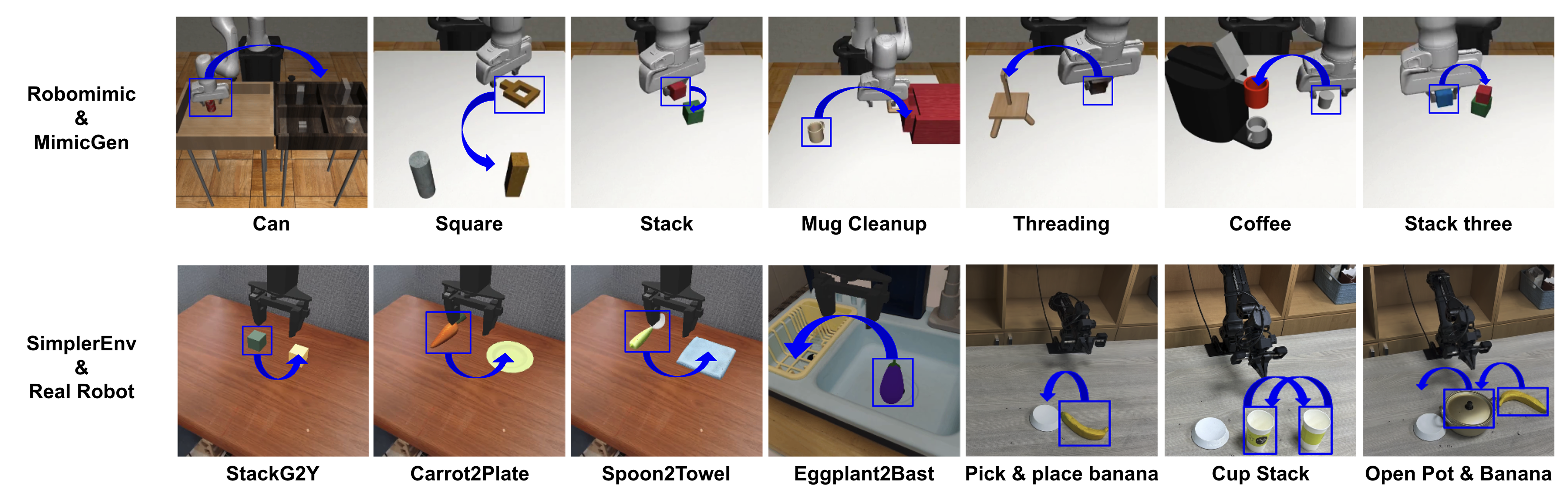}
\vspace{-0.5em}
\caption{\textbf{Evaluation tasks.}
We evaluate PoLAR across simulated and real-world tabletop manipulation tasks, including RoboMimic and MimicGen, SimplerEnv-WidowX, and real robot tasks.}
\label{fig:task_suite}
\vspace{-1.5em}
\end{figure*}

\subsection{PoLAR: Radially Structured Latent Action Pretraining}

We follow the latent action learning pipeline generally used in prior work~\citep{bruce2024geniegenerativeinteractiveenvironments,nikulin2025latent,jeong2026learningactrobustlyviewinvariant,ye2024latent,bu2025univla}.
Given an observation pair $(o_t,o_{t+\ell})$, a visual encoder computes $s_t=f_\xi(o_t)$ and $s_{t+\ell}=f_\xi(o_{t+\ell})$.
An inverse dynamics model (IDM) predicts a continuous latent action $z_{t,\ell}=E_\theta(s_t,s_{t+\ell})$, optionally quantized as $\bar{z}_{t,\ell}=Q(z_{t,\ell})$ for discrete latents, and a forward dynamics model (FDM) reconstructs the future feature from $(s_t,z_{t,\ell})$ in the continuous setting or $(s_t,\bar{z}_{t,\ell})$ in the discrete setting.

PoLAR imposes radial structure on this latent action space: radius is encouraged to encode observed transition extent, reducing the burden on direction to encode extent and leaving direction to capture transition mode.
Rather than directly supervising directional similarity, PoLAR uses temporal ordering as weak radial supervision.
For each start observation $o_t$, we sample $o_{t+j}$ and $o_{t+k}$ with $0<j<k\le K_{\max}$, where $K_{\max}$ is dataset-specific due to differing frame rates, and compute
\[
    z_{t,0}=E_\theta(s_t,s_t), \quad
    z_{t,j}=E_\theta(s_t,s_{t+j}), \quad
    z_{t,k}=E_\theta(s_t,s_{t+k}).
\]
The self pair $(s_t,s_t)$ defines a no-change anchor $z_{t,0}$ for the same starting observation.
The base latent action objective ($\mathcal{L}_{\mathrm{LAM}}$) reconstructs both future features, $s_{t+j}$ and $s_{t+k}$, from $(s_t,\tilde{z}_{t,j})$ and $(s_t,\tilde{z}_{t,k})$, where $\tilde{z}=z$ for continuous latents and $\tilde{z}=\bar{z}$ for discrete latents; in the discrete case, it also includes the codebook and commitment losses.

\noindent\textbf{Hyperbolic radial geometry.}
PoLAR keeps the IDM output $z$ as a Euclidean vector in the tangent space at the origin.
For radial losses, $z$ is lifted to the Poincar\'e ball with curvature $-c$:
\[
    \Phi_c(z)=\exp_0^c(z), \qquad
    \exp_0^c(v)
    =
    \tanh(\sqrt{c}\lVert v\rVert)
    \frac{v}{\sqrt{c}\lVert v\rVert},
\]
with the value at $v=0$ defined by continuity.
The hyperbolic lift is used only to compute radial losses; quantization and FDM decoding use the original tangent-coordinate latent $z$.
We use $c=1$ by default.
Radius and pairwise distance are defined as
\[
    r(z) = d_{\mathbb{H}}(0_{\mathbb{H}},\Phi_c(z)), \qquad
    d(z,z') = d_{\mathbb{H}}(\Phi_c(z),\Phi_c(z')).
\]
The Euclidean ablation uses the same objectives and, when applicable, the same quantizer, but replaces hyperbolic radius and distance with the Euclidean norm and $\ell_2$ distance in tangent coordinates.

\noindent\textbf{Ordering losses.}
PoLAR adds two losses to structure the pre-quantized IDM outputs $z$.
First, the farther transition should lie farther from the local anchor than the intermediate transition:
\[
    \mathcal{L}_{\mathrm{ord}}
    =
    \mathrm{softplus}
    \left(
        d(z_{t,0},z_{t,j})
        -
        d(z_{t,0},z_{t,k})
    \right).
\]
Second, radius should increase with temporal offset:
\[
    \mathcal{L}_{\mathrm{rad}}
    =
    \mathrm{softplus}\left(\alpha j + r(z_{t,0}) - r(z_{t,j})\right)
    +
    \mathrm{softplus}\left(\alpha(k-j) + r(z_{t,j}) - r(z_{t,k})\right),
\]
where $\mathrm{softplus}(x)=\log(1+\exp x)$ is a smooth hinge-like penalty, and $\alpha$ controls the temporal offset margin.
$\mathcal{L}_{\mathrm{rad}}$ orders latent actions by origin-centered radius.
The self-transition anchor $z_{t,0}=E_\theta(s_t,s_t)$ provides this local no-change reference, allowing $\mathcal{L}_{\mathrm{ord}}$ to order $z_{t,j}$ and $z_{t,k}$ by their distances from the same anchor.
Together, $\mathcal{L}_{\mathrm{ord}}$ encourages a start-conditioned ordering of transition extent, while $\mathcal{L}_{\mathrm{rad}}$ expresses this order in the origin-centered radius.
This encourages self-transitions and small transitions to occupy relatively small radii, making latent magnitude reflect transition extent.
Empirically, combining the two losses yields higher downstream success than using either alone, and we examine this result and the learned latent structure in Section~\ref{sec:analysis}.

In the discrete setting, both losses are computed per slot and averaged over valid samples and slots.
The full objective is
\[
    \mathcal{L}
    =
    \mathcal{L}_{\mathrm{LAM}}
    +
    \lambda_{\mathrm{ord}}\mathcal{L}_{\mathrm{ord}}
    +
    \lambda_{\mathrm{rad}}\mathcal{L}_{\mathrm{rad}},
\]
where $\mathcal{L}_{\mathrm{LAM}}$ denotes the base latent action objective. We use $\lambda_{\mathrm{ord}}=1$, $\lambda_{\mathrm{rad}}=0.3$, and $\alpha=0.05$ by default.

\noindent\textbf{Factorized radial and direction tokens.}
For discrete latent actions, we replace the flat VQ codebook with a factorized radial-direction codebook in the same tangent-coordinate representation.
The codebook contains ordered radii $\{\rho_a\}_{a=1}^{R}$ and normalized directions $\{u_b\}_{b=1}^{D}$.
Given the pre-quantized continuous latent action $z_{t,\ell}$ with $C$ latent slots, the quantizer selects one shared radial index $a_{t,\ell}$ from the mean norm across its $C$ slots and one direction index $b_{t,\ell,m}$ for each slot $m$ from the normalized direction of slot $z_{t,\ell,m}$:
\[
    \bar{z}_{t,\ell,m} = \rho_{a_{t,\ell}} u_{b_{t,\ell,m}},
    \qquad m=1,\ldots,C.
\]
The resulting vectors $\bar{z}_{t,\ell,m}$ replace the flat VQ embeddings in the base latent action objective. The direction codebook is shared across all radii.

\begin{figure*}[t]
    \centering
    \includegraphics[width=0.80\textwidth]{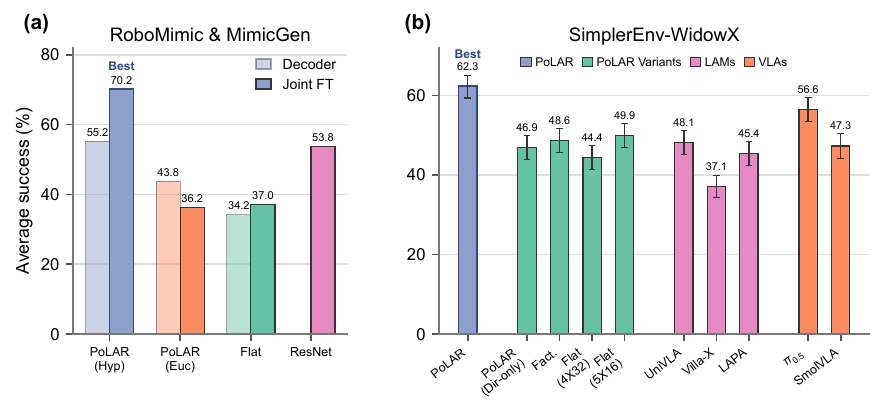}
    \vspace{-1.5em}
    \caption{\textbf{Simulation results.}
(a) PoLAR improves continuous latent action conditioned diffusion policies on RoboMimic and MimicGen.
(b) PoLAR with VLA shows the highest average success rate among baselines on SimplerEnv-WidowX including pretrained latent action models and pretrained VLAs. Error bars in (b): 95\% CIs.}
    \label{fig:sim_sr}
    \vspace{-1.5em}
\end{figure*}

\subsection{Policy Learning from Relabeled Latent Actions}

After pretraining, the IDM relabels action-labeled demonstrations with latent actions for a fixed temporal offset $h$, separate from the randomly sampled offsets $(j,k)$ used for pretraining.
For each transition $(o_t,o_{t+h})$, the IDM returns either a continuous latent action or, in the discrete setting, a radial-direction token sequence $\tau_{t,h}=(a_{t,h}, b_{t,h,1}, \ldots, b_{t,h,C})$.
Downstream control consists of a latent policy and a low-level action decoder.
The latent policy predicts the relabeled latent action from execution-time context (e.g., image, proprioception, language instruction), and the action module maps the latent policy outputs to low-level robot action chunks.
\section{Experimental Results}
\subsection{Experimental Setup}
We evaluate PoLAR across in-task and large-scale pretraining, continuous and discrete latent actions, diffusion policies~\citep{chi2023diffusionpolicy} and VLAs, and simulated and real-world control. Additional experimental details are provided in Appendices~\ref{app:implementation} and~\ref{app:dataset_evaluation}.

\noindent\textbf{In-task pretraining and diffusion policy fine-tuning.}
We evaluate five tasks here: Can, Square, Stack, Mug Cleanup, and Threading (Fig.~\ref{fig:task_suite}), from RoboMimic~\citep{robomimic2021} and MimicGen~\citep{mandlekar2023mimicgen}.
We first train a continuous PoLAR on the task demonstrations, then use the pretrained IDM to relabel the same demonstrations with latent actions.
Following latent action conditioned diffusion policy pipelines in~\citep{jeong2026learningactrobustlyviewinvariant}, a latent policy predicts the relabeled latent action from the current image frame, and a diffusion policy predicts low-level action sequences conditioned on the predicted latent action and proprioception.
We evaluate decoder-only fine-tuning, where the latent policy is frozen and only the diffusion policy is trained, and joint fine-tuning, where both modules are updated.
For each task, we evaluate success over 100 rollout episodes.

\noindent\textbf{Large-scale pretraining and VLA fine-tuning.}
For VLA experiments, we follow the UniVLA-style pipeline~\citep{bu2025univla}.
We train a DINOv2-based~\citep{oquab2024dinov2learningrobustvisual} PoLAR tokenizer in patch-feature space, then train a Prismatic-7B~\citep{karamcheti2024prismaticvlmsinvestigatingdesign} latent VLA policy on BridgeData V2~\citep{walke2023bridgedata}.
The discrete PoLAR interface uses one radial token and four direction tokens from a 16-radius/16-direction factorized codebook.
The pretrained latent VLA is fine-tuned on downstream action-labeled demonstrations with a lightweight multi-head attention pooling action decoder, which pools VLA visual patch states and latent action token hidden states before predicting low-level action chunks.
For SimplerEnv-WidowX~\citep{li24simpler} (four tasks; Fig.~\ref{fig:task_suite}), we fine-tune on 50 successful episodes per task and evaluate on 240 held-out episodes per task that do not overlap with the fine-tuning demonstrations, following the UniVLA execution protocol.
For real-world experiments, we evaluate three tasks with 30 trials per task on WidowX SoloAI robot platforms (Fig.~\ref{fig:task_suite}).
Each BridgeData V2-pretrained latent VLA is fine-tuned for each single task on demonstrations collected on the same platforms.
For sequential tasks, we report success after each required stage as well as final task success.

\begin{figure*}[t]
    \centering
    \includegraphics[width=0.95\textwidth]{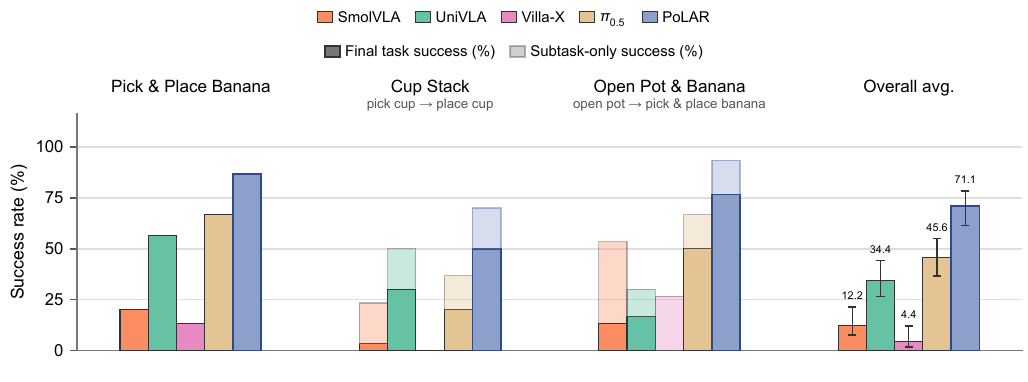}
    
    \caption{\textbf{Real-world results.}
PoLAR with VLA achieves the highest success rates across three real-robot tasks. Error bars: 95\% CIs.}
    \label{fig:real_world_success}
    \vspace{-1.0em}
\end{figure*}

\noindent\textbf{Baselines and controls.}
For in-task diffusion policy experiments, all latent action variants differ only in the radial losses and geometry.
In joint fine-tuning, we also include a pretrained ResNet18~\citep{he2015deepresiduallearningimage} encoder jointly fine-tuned with the same diffusion policy as a non-latent-action baseline. The ResNet18 baseline uses the same batch size and total number of
in-task training updates as the latent-action pipeline.
For VLA experiments, we compare PoLAR ablations, latent action baselines, and pretrained VLA references using the same downstream data and optimization budget, defined as batch size and number of training steps.
In VLA ablations, \emph{dir-only} removes the radial token from the VLA target after PoLAR tokenizer pretraining, \emph{Fact.} uses the radial-direction codebook without radial supervision, and \emph{Flat} uses matched-capacity unfactorized tokenizers with either five 16-way tokens or four 32-way tokens.
UniVLA~\citep{bu2025univla} is the closest data-matched latent action baseline: we use matched training budgets for latent action pretraining, Prismatic-7B VLA training, and downstream fine-tuning.
Villa-X~\citep{chen2025villa0x0} uses its released latent action model, pretrained on a larger mixture including Open-X-Embodiment  and Ego4D~\citep{open_x_embodiment_rt_x_2023,grauman2022ego4dworld3000hours}; we then run Prismatic-7B VLA training and downstream fine-tuning under the same data and optimization budget as PoLAR.
LAPA~\citep{ye2024latent} starts from the released BridgeData V2 checkpoint and is fine-tuned with its original downstream protocol under the same downstream data and optimization budget as PoLAR.
$\pi_{0.5}$~\citep{intelligence2025pi05visionlanguageactionmodelopenworld} and SmolVLA~\citep{shukor2025smolvlavisionlanguageactionmodelaffordable} use LeRobot-provided base checkpoints~\citep{cadene2026lerobotopensourcelibraryendtoend} and are fine-tuned under the same downstream data and optimization budget as PoLAR.
We report them as external VLA references because their base checkpoints are pretrained outside our matched BridgeData V2 pipeline. All experiments use the same fixed top-view camera observations within each dataset.

\begin{figure*}[t]
    \centering
    \includegraphics[width=0.95\textwidth]{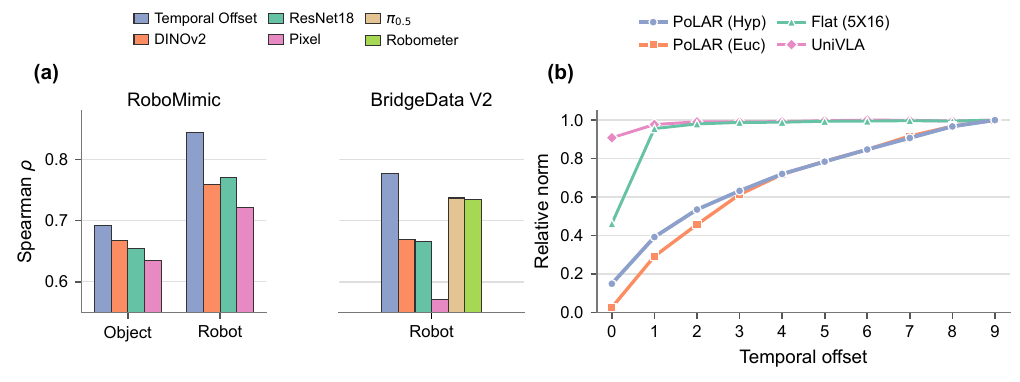}
    \vspace{-1.0em}
    \caption{\textbf{Temporal offset as proxy for radial supervision.}
(a) Temporal offset is an effective proxy for object and robot state change.
(b) PoLAR radii increase with temporal offset, while flat baselines remain nearly constant.}
    \label{fig:temp_offset}
    \vspace{-1.5em}
\end{figure*}

\subsection{Results}

\noindent\textbf{RoboMimic \& MimicGen.}
Fig.~\ref{fig:sim_sr} (a) evaluates diffusion policies conditioned on continuous latent actions, with latent action pretraining and policy fine-tuning performed on RoboMimic and MimicGen tasks.
Across both decoder-only and joint fine-tuning, PoLAR outperforms the \emph{Flat} and the Euclidean variant.
It also outperforms a pretrained ResNet18~\citep{he2015deepresiduallearningimage} encoder jointly fine-tuned with the policy, supporting the value of radial structure in latent action pretraining over generic visual encoder pretraining.

\noindent\textbf{SimplerEnv.}
Fig.~\ref{fig:sim_sr} (b) tests whether PoLAR remains effective through the full VLA pipeline: BridgeData V2 pretraining, discrete latent action tokenization, and downstream fine-tuning on SimplerEnv-WidowX. 
PoLAR achieves the highest average success among all compared methods, including PoLAR ablations, latent action baselines, and pretrained VLA references.
It reaches $62.3\%$ versus $56.6\%$ for $\pi_{0.5}$ over 240 held-out episodes per task (two-sided Fisher's exact test pooled across tasks, $p=0.012$).
Notably, PoLAR uses only BridgeData V2 for latent action and VLA pretraining in our matched pipeline, yet outperforms baselines such as $\pi_{0.5}$ and Villa-X whose released checkpoints use broader pretraining mixtures.
PoLAR outperforms \emph{dir-only}, suggesting that the radial token contributes beyond direction tokens alone.
It also improves over \emph{Fact.}, which keeps the radial-direction codebook without radial supervision, and over matched-capacity \emph{Flat} tokenizers.
These comparisons suggest that PoLAR's gains come from combining radial supervision with radial-direction factorization, rather than from factorization or token count alone.

\noindent\textbf{Real robot.}
Fig.~\ref{fig:real_world_success} evaluates PoLAR on real-world robot manipulation tasks.
PoLAR achieves the highest final success rate on all three tasks and the best overall average, outperforming $\pi_{0.5}$ as well as UniVLA, SmolVLA, and Villa-X.
With 30 trials per task, average final success is $71.1\%$ for PoLAR versus $45.6\%$ for $\pi_{0.5}$ (two-sided Fisher's exact test pooled across tasks, $p<0.001$).
The subtask breakdown suggests that the gains extend beyond early-stage grasping or reaching, supporting the usefulness of radial latent action structure in real-world control.

\section{Analysis}
\label{sec:analysis}


\subsection{Radial Structure}

\noindent\textbf{Temporal offset as a proxy for transition extent.}
PoLAR uses temporal ordering as weak supervision for transition extent, so we first test whether this signal tracks state change.
Fig.~\ref{fig:temp_offset} (a) compares temporal offset with visual feature distances as proxies for state distance in RoboMimic and BridgeData V2.
Across all datasets, temporal offset has the strongest Spearman correlation with both object and robot state distances, outperforming pretrained DINOv2~\citep{oquab2024dinov2learningrobustvisual}, ResNet18~\citep{he2015deepresiduallearningimage}, and pixel distances in all reported settings. Additional comparisons with $\pi_{0.5}$ and Robometer~\citep{liang2026robometerscalinggeneralpurposerobotic} features
on BridgeData V2 show the same trend.
This supports temporal ordering as an effective signal for observed transition extent.

\begin{figure*}[t]
    \centering
    \begin{minipage}[c]{0.60\textwidth}
        \centering
        \includegraphics[width=\linewidth]{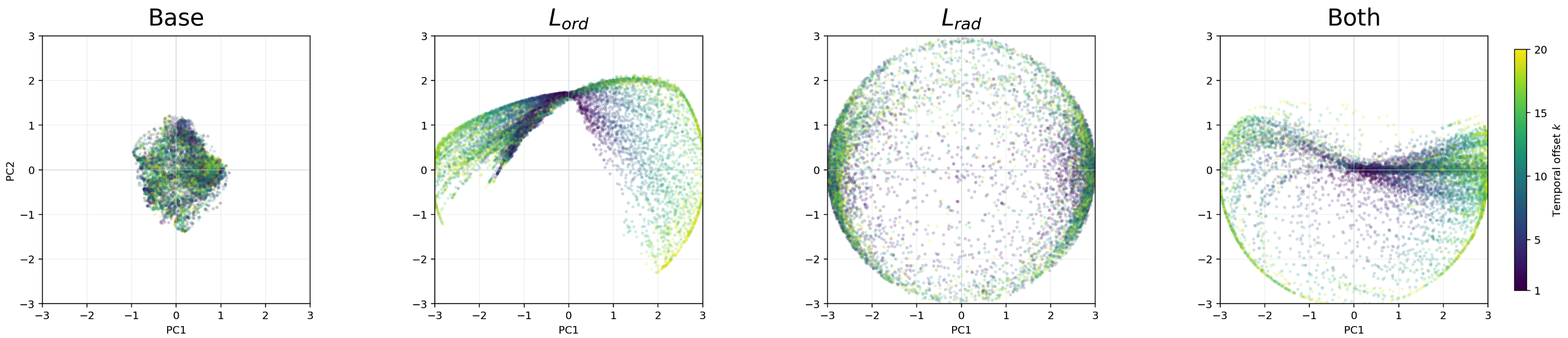}
        \caption{\textbf{Effect of the ordering losses on Can.}
        PCA of latent actions, colored by temporal offset. Lighter colors indicate larger offset.}
        \label{fig:pca}
    \end{minipage}
    \hfill
    \begin{minipage}[c]{0.37\textwidth}
        \centering
        \captionof{table}{\textbf{Action informativeness.}
        MI estimates and probe $R^2$ from latent actions to ground-truth action chunks.}
        \label{tab:action_informativeness}
        \resizebox{\linewidth}{!}{%
        \begin{tabular}{lccc}
            \toprule
            \textbf{Model} & \textbf{BA $\uparrow$}
            & \textbf{InfoNCE $\uparrow$} & \textbf{$R^2$ $\uparrow$} \\
            \midrule
            \rowcolor{blue!20}
            PoLAR (Hyp) & \textbf{8.552} & \textbf{0.672} & \textbf{0.184} \\
            PoLAR (Euc) & 6.470 & 0.560 & 0.160 \\
            \emph{Fact.} & 5.266 & -0.042 & 0.098 \\
            \emph{Flat ($5\times16$)} & 7.496 & -0.106 & 0.160 \\
            \emph{Flat ($4\times32$)} & 6.441 & 0.138 & 0.107 \\
            \emph{Flat ($4\times16$)} & 6.290 & 0.026 & 0.094 \\
            UniVLA~\citep{bu2025univla} & 2.720 & -1.186 & 0.035 \\
            \bottomrule
        \end{tabular}%
        }
    \end{minipage}
    \vspace{-2.0em}
\end{figure*}

\noindent\textbf{Radius tracks temporal offset.}
We next test whether PoLAR turns this signal into an ordered radial coordinate.
Fig.~\ref{fig:temp_offset} (b) shows that PoLAR produces a gradual radial progression as temporal offset increases.
In contrast, flat latent action baselines remain nearly constant, suggesting that transition extent is not naturally learned along radius without radial supervision.

\begin{figure*}[t]
    \centering
    \includegraphics[width=0.80\textwidth]{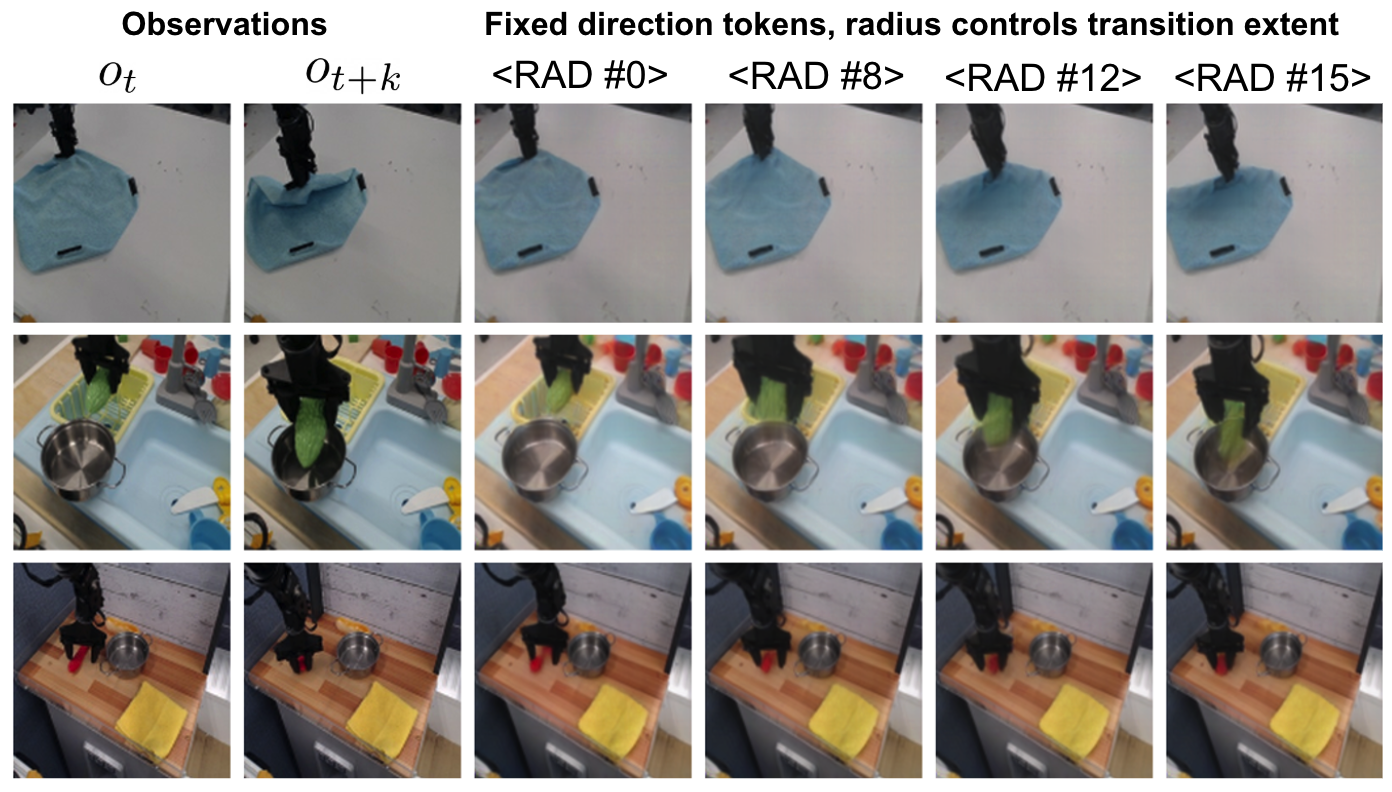}
    \vspace{-0.5em}
    \caption{\textbf{Radius controls transition extent.}
With direction tokens fixed, increasing the radial token produces progressively larger visual transitions.}

    \label{fig:radius_sweep_decoder}
    \vspace{-1.5em}
\end{figure*}

\noindent\textbf{Radius and direction play distinct roles.}
Figures~\ref{fig:teaser} and~\ref{fig:radius_sweep_decoder} provide a qualitative intervention on the learned structure.
We fix the direction of a latent action and sweep only the radial code.
We visualize each swept latent by applying the pretrained FDM and decoding the predicted DINOv2 feature with a separately trained VQ-VAE pixel decoder~\citep{oord2018neuraldiscreterepresentationlearning}.
As radius increases, the decoded visual transition becomes larger while preserving the transition mode.
This behavior is consistent with the intended factorization: radius represents transition extent, while direction represents transition mode. Additional qualitative examples and decoder details are provided in Appendices~\ref{app:radius_sweep_more} and~\ref{app:radius_sweep_protocol}.

\noindent\textbf{Both radial losses matter.}
Table~\ref{tab:polar_ablation} ablates PoLAR components and radial-margin hyperparameters on RoboMimic Can and Square.
Removing either $\mathcal{L}_{\mathrm{ord}}$ or $\mathcal{L}_{\mathrm{rad}}$ weakens downstream policy performance under both fine-tuning settings, showing that the two losses are complementary in practice.
Fig.~\ref{fig:pca} shows that $\mathcal{L}_{\mathrm{ord}}$ alone orders transitions by temporal offset, but short-offset transitions remain away from the plotted origin.
Using both losses brings these transitions closer to the plotted origin while larger-offset transitions spread outward.
On Can, the latent norm for two identical images is $59.4\%$ of that for images 20 steps apart with $\mathcal{L}_{\mathrm{ord}}$ alone.
Using both losses reduces this ratio to $13.7\%$.
The correlation between temporal offset and norm remains high in both cases (Table~\ref{tab:polar_ablation} (a)).
Together with the norm trend in Fig.~\ref{fig:temp_offset} (b), these results support the intended ordering around the origin: little or no change receives a small relative norm.
Ideally, latent actions should capture changes between observations rather than static scene information, but a reconstruction objective alone does not guarantee this.
Our results suggest that PoLAR moves closer to this goal, learning latent actions more focused on what changes between observations.

\begin{table*}[t]
    \centering
    \caption{\textbf{Ablations and radial structure analysis.}
    Success (\%): mean$\pm$SD over three seeds; Dec. and FT denote decoder-only and joint fine-tuning.
    In (a), self-gap is $1-\|z_0\|/\|z_{20}\|$ and $\rho=\rho(k,\|z_k\|)$ on Can,
    where $z_k$ is the latent action from $(o_t,o_{t+k})$.
    The colored rows indicate the final setting.}
    \label{tab:polar_ablation}
    \label{tab:radial_organization}
    \label{tab:ablation_can}
    \label{tab:curvature}
    \small
    \setlength{\tabcolsep}{4pt}
    \renewcommand{\arraystretch}{1.05}
    \textbf{(a) Loss components and radial structure}\par\vspace{3pt}
    \begin{tabular}{lcccccc}
        \toprule
        & \multicolumn{2}{c}{\textbf{Radial structure (Can)}}
        & \multicolumn{2}{c}{\textbf{Can success (\%)}}
        & \multicolumn{2}{c}{\textbf{Square success (\%)}} \\
        \cmidrule(lr){2-3} \cmidrule(lr){4-5} \cmidrule(lr){6-7}
        \textbf{Variant} & \textbf{Self-gap $\uparrow$} & $\boldsymbol{\rho}\uparrow$
        & \textbf{Dec.} & \textbf{FT} & \textbf{Dec.} & \textbf{FT} \\
        \midrule
        Base & .002 & .202
        & $66.7\pm1.2$ & $59.3\pm4.2$ & $58.0\pm7.2$ & $69.3\pm2.3$ \\
        $+\mathcal{L}_{\mathrm{ord}}$ & .406 & \textbf{.976}
        & $17.3\pm9.2$ & $54.0\pm11.1$ & $54.7\pm3.1$ & $71.3\pm4.6$ \\
        $+\mathcal{L}_{\mathrm{rad}}$ & .179 & .466
        & $44.0\pm15.1$ & $39.3\pm15.5$ & $35.3\pm13.3$ & $52.7\pm5.0$ \\
        \rowcolor{blue!20}
        +Both & \textbf{.863} & .968
        & $\mathbf{92.7\pm6.1}$ & $\mathbf{80.0\pm14.0}$
        & $\mathbf{71.3\pm7.0}$ & $\mathbf{73.3\pm4.2}$ \\
        \bottomrule
    \end{tabular}
    \par\medskip
    \begin{minipage}[t]{0.73\linewidth}
        \centering
        \textbf{(b) Hyperparameter $(\lambda_{\mathrm{rad}},\alpha)$}\par\vspace{3pt}
        \setlength{\tabcolsep}{3pt}
        \resizebox{\linewidth}{!}{%
        \begin{tabular}{lcccc}
            \toprule
            & \multicolumn{2}{c}{\textbf{Can success (\%)}}
            & \multicolumn{2}{c}{\textbf{Square success (\%)}} \\
            \cmidrule(lr){2-3} \cmidrule(lr){4-5}
            $(\lambda_{\mathrm{rad}},\alpha)$
            & \textbf{Dec.} & \textbf{FT} & \textbf{Dec.} & \textbf{FT} \\
            \midrule
            $(0.10,0.05)$
            & $73.7\pm11.1$ & $59.3\pm6.1$ & $62.0\pm9.8$ & $70.3\pm6.8$ \\
            $(0.30,0)$
            & $70.3\pm11.1$ & $54.7\pm18.9$ & $70.0\pm6.0$ & $67.7\pm5.7$ \\
            $(0.30,0.025)$
            & $65.3\pm1.2$ & $48.0\pm21.8$
            & $\mathbf{73.3\pm8.1}$ & $\mathbf{74.7\pm11.4}$ \\
            \rowcolor{blue!20}
            $(0.30,0.05)$
            & $\mathbf{92.7\pm6.1}$ & $\mathbf{80.0\pm14.0}$
            & $71.3\pm7.0$ & $73.3\pm4.2$ \\
            $(0.30,0.10)$
            & $77.7\pm5.1$ & $65.7\pm9.0$ & $70.0\pm7.9$ & $73.0\pm1.7$ \\
            $(1.0,0.05)$
            & $73.7\pm8.4$ & $71.3\pm4.2$ & $62.0\pm21.0$ & $65.3\pm6.1$ \\
            \bottomrule
        \end{tabular}%
        }
    \end{minipage}
    \hfill
    \begin{minipage}[t]{0.24\linewidth}
        \centering
        \textbf{(c) Curvature}\par\vspace{3pt}
        \setlength{\tabcolsep}{3pt}
        \resizebox{\linewidth}{!}{%
        \begin{tabular}{lc}
            \toprule
            $c$ & \textbf{Can FT (\%)} \\
            \midrule
            $0$ (Euc) & $34.3\pm6.8$ \\
            $0.1$ & $36.7\pm9.2$ \\
            $0.25$ & $62.7\pm18.0$ \\
            $0.5$ & $65.7\pm16.1$ \\
            \rowcolor{blue!20}
            $1$ & $\mathbf{80.0\pm14.0}$ \\
            $2$ & $73.7\pm9.1$ \\
            $4$ & $72.3\pm14.0$ \\
            \bottomrule
        \end{tabular}%
        }
    \end{minipage}
    \par\vspace{-8pt}
\end{table*}

\subsection{Advantages of PoLAR for Robot Policy Learning}

Having verified the intended radial organization, we next analyze three policy-relevant benefits of this structure.
Additional details are provided in Appendices~\ref{app:action_informativeness}, \ref{app:wrong_token_error}, and~\ref{app:multi_horizon}.

\noindent\textbf{Action informativeness.}
We quantify how much information learned latent actions contain about ground-truth robot action chunks on BridgeData V2.
We estimate mutual information using Barber--Agakov (BA)~\citep{barber2003} and InfoNCE~\citep{oord2019representationlearningcontrastivepredictive} variational bounds~\citep{poole2019variationalboundsmutualinformation}, and train attentive probes~\citep{bardes2024revisitingfeaturepredictionlearning} from latent actions to action chunks.
We report probe $R^2$ ($=1-\mathrm{SSE}/\mathrm{SST}$), the fraction of action variance explained by the probe predictions; the attentive pooling layer aggregates over latent action tokens so representations with different token counts can be compared.
Ground-truth actions are used only for this diagnostic, not for latent action pretraining.
Table~\ref{tab:action_informativeness} shows that hyperbolic PoLAR achieves the highest mutual information estimates and probe $R^2$.
Together with the downstream gains over \emph{Fact.} in Fig.~\ref{fig:sim_sr} (b), this suggests that codebook factorization alone is insufficient; radial structure in hyperbolic space helps retain action-related information in the learned latent actions.

\noindent\textbf{Robustness to prediction errors.}
We test whether mispredicted latent action tokens remain close to the target and decode to small action errors.
Hyperbolic PoLAR yields lower wrong-token latent error (normalized) than \emph{Flat} and UniVLA (0.311 vs. 0.447 and 0.689), and the same trend holds after decoding to actions (0.143 vs. 0.238 and 0.191).
This is consistent with the motivation of PoLAR: if extent and mode are represented separately, some token errors can remain close to the intended transition instead of mapping to an unrelated flat code.

\noindent\textbf{Multi-horizon latent policy training.}
The main experiments use single-horizon latent action prediction, where the latent policy predicts the latent action for one fixed future offset.
We also analyze a multi-horizon variant that predicts concatenated latent actions across multiple future offsets from the same observation window.
PoLAR benefits from multi-horizon latent policy training on Coffee and Stack Three (Fig.~\ref{fig:task_suite}) in diffusion policy (+36.0 and +60.0 points), and also improves the SimplerEnv average in VLA (+4.2 points); \emph{Flat} shows no gain (-4.0, 0.0, and 0.0 points).
On SimplerEnv, PoLAR also shows higher cross-horizon gradient cosine similarity than \emph{Flat} (0.486 vs. 0.305), suggesting that when different horizons share direction structure and differ mainly in radius, their training targets induce less conflicting gradients.

\subsection{Why Hyperbolic Geometry?}

Both hyperbolic and Euclidean PoLAR learn radii that increase with temporal offset (Fig.~\ref{fig:temp_offset} (b)), so the key difference is not whether radius can represent transition extent.
The Euclidean variant has higher cosine similarity between directions at adjacent temporal offsets than hyperbolic PoLAR (0.974 vs. 0.908), but lower direction-only probe $R^2$ after removing radius (0.088 vs. 0.208).
This suggests that Euclidean radial supervision relies more on norm expansion: directions change little across temporal offsets, yet carry less action-predictive information.
In contrast, hyperbolic geometry better supports PoLAR's intended structure: as radius tracks transition extent, the expanding angular capacity at larger radii helps direction retain information for distinguishing transition modes.
This helps explain why hyperbolic PoLAR is more action-informative overall in Table~\ref{tab:action_informativeness}.
The curvature ablation shows that success drops substantially as $c$ approaches the Euclidean limit, while increasing $c$ beyond $1$ leads to a more modest decline (Table~\ref{tab:polar_ablation} (c)).
This suggests that PoLAR benefits from hyperbolic geometry without requiring a narrowly tuned curvature within the tested range of $c=1$ to $4$.

\section{Limitations}

PoLAR infers latent action structure from visual observation pairs, and our experiments use a fixed third-person, top-view camera.
Extending PoLAR to multi-view settings, including both third-person and wrist cameras, is a natural direction for future work.
Also, PoLAR uses temporal ordering as weak supervision for transition extent.
This assumption is well suited to goal-directed demonstrations, but may break down under cyclic behavior, pauses, recovery motions, or repeated back-and-forth moves.
Performance is also sensitive to the radial weight and margin, with the best hyperparameter setting varying across tasks.
\section{Conclusion}
We introduced PoLAR, a latent action learning framework that uses temporal ordering to impose hyperbolic radial structure on latent actions, encouraging radius to represent transition extent and direction to retain transition mode.
Across various policies, simulation benchmarks, and real-world robot experiments, PoLAR consistently improves downstream policy performance as a pretraining method.
Our analyses further link these gains to better organized and more action-informative latent actions.
Together, these results suggest that latent action geometry is a useful design choice for robot policy learning.


\clearpage



\noindent\textbf{Acknowledgements.}
This work was supported by the Institute of Information \& communications Technology Planning \& Evaluation (IITP) under the Leading Generative AI Human Resources Development (IITP-2026-RS-2024-00397085) grant funded by the Korea government (MSIT); the National Research Foundation of Korea (NRF) grant funded by the Korea government (MSIT) (No. RS-2024-00345809, Research on AI Robustness Against Distribution Shift in Real-World Scenarios); the Advanced GPU Utilization Support Program funded by the Government of the Republic of Korea (Ministry of Science and ICT); and the NVIDIA Academic Grant Program using NVIDIA H100 GPUs.

\bibliography{example}  

\clearpage
\appendix
\renewcommand{\thefigure}{A.\arabic{figure}}
\renewcommand{\thetable}{A.\arabic{table}}
\setcounter{figure}{0}
\setcounter{table}{0}

\section*{Appendix}
\section{Implementation Details}
\label{app:implementation}
\subsection{Latent Action Pretraining}

\paragraph{PoLAR pretraining on RoboMimic \& MimicGen.}
Table~\ref{tab:app_mimic_polar_hparams} summarizes the hyperparameters used for continuous PoLAR pretraining on RoboMimic~\citep{robomimic2021} and MimicGen~\citep{mandlekar2023mimicgen}.
This stage learns a continuous latent action model directly on task demonstrations.
We encode the current and target images with a 3-stage residual CNN and train latent actions in the resulting visual feature space.
Unlike BridgeData V2 pretraining, where we use a frozen DINOv2 encoder, the RoboMimic and MimicGen experiments jointly train the visual encoder with the IDM and FDM.
The PoLAR objective adds the ordering and radial losses, with $\lambda_{\mathrm{ord}}=1$, $\lambda_{\mathrm{rad}}=0.3$, and margin $\alpha=0.05$.
Unless otherwise noted, all continuous PoLAR ablations use the same visual encoder, IDM/FDM architecture, data, optimizer, augmentation, and training schedule as Table~\ref{tab:app_mimic_polar_hparams}; \emph{Flat} removes radial supervision, and PoLAR (Euc) uses the same radial losses in Euclidean space.
  
\paragraph{PoLAR pretraining on BridgeData V2.}
Table~\ref{tab:app_bridge_tokenizer_hparams} summarizes the hyperparameters used for PoLAR pretraining on BridgeData V2~\citep{walke2023bridgedata}.
We use a frozen DINOv2 ViT-B/14 with registers~\citep{oquab2024dinov2learningrobustvisual} and train the tokenizer in DINO patch-feature space.
The IDM is a spatiotemporal transformer that encodes the current and target patch features and outputs four continuous latent code vectors for each transition.
The FDM is a spatial transformer that reconstructs target DINO patch features from the current patch features and the quantized latent codes.

For discrete PoLAR, we replace a flat VQ codebook with a radial-direction VQ interface.
Let $e\in\mathbb{R}^{4\times d}$ denote the four continuous latent code vectors produced by the IDM.
We select one shared radial code by quantizing the average norm of these latent code vectors, and select one direction code per latent code by matching its normalized direction to the nearest direction code.
Thus each discrete latent action is represented by one radial token and four direction tokens.
The latent code vocabulary is split into 16 radial IDs and 16 direction IDs.

We train the quantizer with a straight-through estimator.
Given the quantized code vectors $z$, the VQ loss is
\[
    \mathcal{L}_{\mathrm{VQ}}
    =
    \lVert \mathrm{sg}(e)-z\rVert_2^2
    +
    \beta \lVert e-\mathrm{sg}(z)\rVert_2^2,
\]
where $\mathrm{sg}(\cdot)$ denotes stop-gradient and $\beta=0.25$ is the commitment weight.
The tokenizer is trained with DINO feature reconstruction, the VQ loss, and the PoLAR radial losses.
Unless otherwise noted, ablations use the same DINO feature space, IDM/FDM architecture, data, and optimization settings as Table~\ref{tab:app_bridge_tokenizer_hparams}; \emph{Fact.} keeps the radial-direction codebook but removes PoLAR radial supervision, while \emph{Flat} replaces the radial-direction quantizer with matched-capacity unfactorized VQ codebooks.

For UniVLA, we keep its original fixed-horizon 4-token, 16-code latent action interface and single-target transition reconstruction.
Because the PoLAR tokenizer reconstructs two target transitions for each sampled $(j,k)$ pair, we train each UniVLA latent-action stage for 60k steps to match the FDM reconstruction count of the 30k-step PoLAR tokenizer.

\begin{table}[t]
\centering
\caption{\textbf{RoboMimic and MimicGen continuous PoLAR hyperparameters.}}
\label{tab:app_mimic_polar_hparams}
\begin{tabular}{lc}
\toprule
\textbf{Name} & \textbf{Value} \\
\midrule
Visual encoder & 3-stage residual CNN \\
Image resolution & $64\times64$ \\
$K_{\max}$ & 20 \\
Encoder blocks & 3 stride-2 conv blocks, 2 residual blocks each \\
Encoder channels & $[96,192,192]$ \\
Encoder output & $192\times8\times8$ feature map with LayerNorm \\
Latent action dimension & 256 \\
IDM / FDM width & 256 \\
IDM / FDM architecture & 3-block MLP \\
PoLAR loss weights & $\lambda_{\mathrm{ord}}=1$, $\lambda_{\mathrm{rad}}=0.3$ \\
Radial margin & $\alpha=0.05$ \\
Curvature & $c=1$ \\
Optimizer & AdamW \\
Learning rate / weight decay & $10^{-4}$ / $10^{-6}$ \\
Batch size & 256 \\
Scheduler & cosine decay with warmup \\
Training steps & 5,000 (Can), 15,000 (others) \\
\bottomrule
\end{tabular}
\end{table}
  
\begin{table}[t]
\centering
\caption{\textbf{BridgeData V2 PoLAR tokenizer hyperparameters.}}
\label{tab:app_bridge_tokenizer_hparams}
\begin{tabular}{lc}
\toprule
\textbf{Name} & \textbf{Value} \\
\midrule
Visual encoder & frozen DINOv2 ViT-B/14-reg \\
Image resolution & $224\times224$ \\
DINO normalization & ImageNet mean/std \\
$K_{\max}$ & 9 \\
IDM / FDM width & 768 \\
Encoder / decoder blocks & 12 / 12 \\
Attention heads & 12 \\
Code dimension & 128 per code \\
Radius codes & 16 \\
Direction codes & 16 \\
Token layout & 1 radial + 4 direction tokens \\
Token vocabulary & 16 radial + 16 direction IDs \\
VQ loss & $\mathcal{L}_{q} + 0.25\mathcal{L}_{\mathrm{commit}}$ \\
PoLAR loss weights & $\lambda_{\mathrm{ord}}=1$, $\lambda_{\mathrm{rad}}=0.3$ \\
Radial margin & $\alpha=0.05$ \\
Curvature & $c=1$ \\
Optimizer & AdamW \\
Learning rate / weight decay & $10^{-4}$ / $10^{-2}$ \\
Global batch size & 512 \\
Training steps & 30,000 \\
Gradient clipping & 0.1 \\
VQ restart interval & 10k steps \\
Image augmentation & random crop / color jitter \\
\bottomrule
\end{tabular}
\end{table}

\subsection{Latent Policies}

Tables~\ref{tab:app_stage2_hparams} and~\ref{tab:app_vla_pretrain_hparams} summarize the latent policy training hyperparameters.
After latent action pretraining, we freeze the pretrained IDM and use it to relabel action-labeled demonstrations at a fixed policy horizon.
The latent policy is then trained to predict these relabeled latent actions from execution-time observations.

\paragraph{Latent Policies for RoboMimic \& MimicGen.} For RoboMimic and MimicGen, we train a continuous latent policy on each task dataset.
The policy takes the current agent-view image as input and predicts the continuous latent action for a fixed horizon $h=20$ using an MSE objective.
This latent policy is later used to condition the low-level diffusion policy.

\paragraph{Latent Policies for BridgeData V2.}
For BridgeData V2, we train a latent VLA policy that observes the current image and language instruction and autoregressively predicts the discrete PoLAR token sequence.
We follow the UniVLA-style action-token interface by adding latent-action special tokens $\{\texttt{<ACT\_0>},\ldots,\texttt{<ACT\_31>}\}$ to the VLA tokenizer.
For PoLAR, $\texttt{<ACT\_0>}$--$\texttt{<ACT\_15>}$ denote radial codes and $\texttt{<ACT\_16>}$--$\texttt{<ACT\_31>}$ denote direction codes.
The prediction target consists of one radial token followed by four direction tokens produced by the frozen PoLAR tokenizer at fixed horizon $h=9$.
The latent VLA is trained with next-token cross entropy over these appended latent-action special tokens, with labels masked on the visual-language prompt tokens.
During autoregressive generation, we apply a slot-wise action-token mask so that the first latent-action position can only emit radial tokens and the remaining positions can only emit direction tokens.
Starting from Prismatic-7B DINO-SigLIP, we update the full VLA parameters and use the resulting checkpoint for downstream fine-tuning.

\begin{table}[t]
  \centering
  \caption{\textbf{RoboMimic and MimicGen continuous latent policy hyperparameters.}}
  \label{tab:app_stage2_hparams}
  \begin{tabular}{lc}
  \toprule
  \textbf{Name} & \textbf{Value} \\
  \midrule
  Visual encoder & 3-stage residual CNN \\
  Encoder blocks & 3 stride-2 conv blocks, 2 residual blocks each \\
  Encoder channels & $[512,1024,1024]$ \\
  Image resolution & $64\times64$ \\
  Temporal offset & fixed $h=20$ \\
  Training steps & 5,000 (Can), 15,000 (others) \\
  Global batch size & 256 \\
  Optimizer & AdamW \\
  Learning rate / weight decay & $10^{-4}$ / $10^{-6}$ \\
  Scheduler & cosine decay with warmup \\
  Image augmentation & random shift / rotation / perspective \\
  \bottomrule
  \end{tabular}
  \end{table}
  
\begin{table}[t]
\centering
\caption{\textbf{BridgeData V2 Latent VLA hyperparameters.}}
\label{tab:app_vla_pretrain_hparams}
\begin{tabular}{lc}
\toprule
\textbf{Name} & \textbf{Value} \\
\midrule
Base VLM & Prismatic-7B DINO-SigLIP~\citep{karamcheti2024prismaticvlmsinvestigatingdesign}\\
Image resolution & $224\times224$ \\
Latent action target & 1 radial + 4 direction tokens \\
Temporal offset & fixed $h=9$ \\
Training steps & 50,000 \\
Global batch size & 256 \\
Optimizer & AdamW \\
Learning rate / weight decay & $2\times10^{-5}$ / 0 \\
Gradient clipping & 1.0 \\
Image augmentation & random crop / color jitter \\
\bottomrule
\end{tabular}
\end{table}

\subsection{Downstream Policy Fine-tuning}
\paragraph{Diffusion policy fine-tuning.}
Table~\ref{tab:app_stage3_hparams} summarizes the low-level diffusion policy used for RoboMimic and MimicGen.
After training the continuous latent policy, we train a conditional 1D diffusion policy to predict low-level 7-DoF action chunks.
The diffusion policy is conditioned on the predicted latent action and proprioception.
It predicts 20-step action chunks.
In the decoder-only setting, the latent policy is frozen and only the diffusion policy is trained; in joint fine-tuning, both the latent policy and diffusion policy are updated.
Because the tasks differ in dataset size and difficulty, we use task-specific diffusion fine-tuning budgets; within each task, all baselines use the same budget.
Table~\ref{tab:app_dp_finetune_steps} lists the budgets used for decoder-only and joint fine-tuning.

\paragraph{VLA action decoder fine-tuning.}
Table~\ref{tab:app_vla_action_decoder_hparams} summarizes the downstream VLA action decoder fine-tuning setup.
Starting from the BridgeData V2-pretrained latent VLA, we attach a lightweight action decoder that maps VLA hidden states to low-level robot action chunks.
The decoder pools final-layer visual patch states and latent action token states with multi-head attention pooling, then predicts normalized 7-DoF action chunks with a linear $\tanh$ head.
During downstream fine-tuning, we update the VLA LoRA adapters and action decoder using action-labeled demonstrations, while keeping the same downstream data and optimization budget across VLA methods.
The objective combines action L1 loss with the autoregressive latent-token cross-entropy loss.

\begin{table}[t]
\centering
\caption{\textbf{Diffusion policy hyperparameters.}}
\label{tab:app_stage3_hparams}
\begin{tabular}{lc}
\toprule
\textbf{Name} & \textbf{Value} \\
\midrule
Proprio encoder & MLP, hidden dim 128, output dim 64 \\
Policy conditioning & predicted latent action + proprio embedding \\
Low-level policy & conditional 1D diffusion decoder \\
Action dimension & 7 \\
Observation horizon & 1 \\
Prediction horizon & 20 \\
Action horizon & 10 \\
Noise scheduler & DDIM \\
Training diffusion steps & 100 \\
Inference diffusion steps & 10 \\
Beta schedule & squared cosine cap v2 \\
UNet diffusion-step embedding dim & 256 \\
UNet down dimensions & $[256,512,1024]$ \\
UNet kernel size / groups & 5 / 8 \\
Optimizer & AdamW \\
Learning rate / weight decay & $10^{-4}$ / $10^{-6}$ \\
Batch size & 256 \\
Scheduler & cosine decay with warmup \\
\bottomrule
\end{tabular}
\end{table}

\begin{table}[t]
\centering
\caption{\textbf{Diffusion policy fine-tuning steps for RoboMimic and MimicGen.}}
\label{tab:app_dp_finetune_steps}
\begin{tabular}{lcc}
\toprule
\textbf{Task} & \textbf{Decoder-only} & \textbf{Joint fine-tuning} \\
\midrule
Can & 5,000 & 2,000 \\
Square & 20,000 & 10,000 \\
Stack & 10,000 & 2,000 \\
Mug Cleanup & 10,000 & 5,000 \\
Threading & 10,000 & 10,000 \\
\bottomrule
\end{tabular}
\end{table}

\begin{table}[t]
\centering
\caption{\textbf{VLA action decoder fine-tuning hyperparameters.}}
\label{tab:app_vla_action_decoder_hparams}
\begin{tabular}{lc}
\toprule
\textbf{Name} & \textbf{Value} \\
\midrule
Fine-tuning method & LoRA + action decoder \\
LoRA rank / dropout & 32 / 0.0 \\
Action decoder input & VLA visual patch states + latent action token states \\
Latent action token states & 1 radial + 4 direction tokens \\
Action decoder pooling & multi-head attention pooling (1 latent, 8 heads) \\
Action decoder hidden dim & 512 \\
Action dimension & 7 \\
Action chunk size & 10 \\
Output head & linear + $\tanh$ \\
Objective & action L1 + latent-token cross entropy \\
Training steps & 10,000 \\
Batch size & 16 \\
Optimizer & AdamW \\
Learning rate / weight decay & $3.5\times10^{-4}$ / $10^{-3}$ \\
Image augmentation & random crop / color jitter \\
\bottomrule
\end{tabular}
\end{table}

\subsection{Compute Cost}

For the continuous diffusion-policy pipeline, we report a representative Square run on a single NVIDIA GeForce RTX 3090.
Continuous PoLAR pretraining used approximately 1.7 GPU-hours, continuous latent policy training used approximately 2.7 GPU-hours, and diffusion policy fine-tuning used approximately 3.0 GPU-hours.

For the discrete VLA pipeline used in the SimplerEnv and real-world experiments, PoLAR tokenizer pretraining used 8 NVIDIA B200 GPUs for approximately 68 GPU-hours, latent VLA training used 8 NVIDIA B200 GPUs for approximately 144 GPU-hours, and downstream fine-tuning used 4 NVIDIA B200 GPUs for approximately 14 GPU-hours.

\section{Analysis Details}

\subsection{Temporal Offset and State Distance}

We audit temporal offset as a weak proxy for physical transition extent by comparing it against low-dimensional state distances.
For RoboMimic, we use Can and Square.
For each dataset, we randomly sample 50 demonstrations, and evaluate offsets $\{1,2,4,8,12,16,20\}$.
State distances are computed as L2 distances after z-scoring the relevant low-dimensional state vectors: object state for \emph{object}, and end-effector pose, gripper position, and joint position for \emph{robot}.
For BridgeData V2, we run the same audit on 49 episodes using offsets $\{1,2,4,8\}$.
BridgeData V2 provides a 7D robot/end-effector proprioceptive state rather than object state, so we use z-scored state L2 distance as the low-dimensional target for \emph{robot}.
For image-based proxies, we use direct endpoint distances between $o_t$ and $o_{t+k}$ from the available RGB observations: agent-view images for RoboMimic and the top-view RGB stream for BridgeData V2.

\subsection{Radius Sweep Visualization Protocol}
\label{app:radius_sweep_protocol}

For the radius-sweep visualization, we use a pretrained BridgeData V2 PoLAR tokenizer and FDM.
For each selected example, we first infer its discrete PoLAR latent action.
We keep the four direction tokens fixed and replace only the shared radial token with different radius indices from the factorized codebook.
Each modified latent action is passed to the pretrained FDM together with the current DINOv2 patch features, producing a predicted future DINOv2 patch-feature map.

To visualize the predicted DINOv2 features in pixel space, we use a separately trained pixel decoder.
The decoder is trained on BridgeData V2 frames for 50k steps with AdamW, batch size 64, learning rate $10^{-4}$, weight decay $10^{-4}$, and gradient clipping at 1.0.
The training loss is a weighted sum of pixel L1 loss, pixel MSE loss, and a DINO feature-cycle loss with weights $1.0$, $0.1$, and $0.2$, respectively.
The decoder is used only for qualitative visualization and is not used for latent action pretraining, VLA training, or downstream policy learning.

\subsection{Action Informativeness}
\label{app:action_informativeness}

We evaluate action informativeness on 4,096 BridgeData V2 samples at horizon $h=9$.
For each sample, we store the latent action produced by the pretrained tokenizer and the corresponding ground-truth 10-step action chunk.
We use the full action sequence as the action target, flattened to a 70-D vector.

For mutual-information diagnostics, we estimate Barber--Agakov~\citep{barber2003} and InfoNCE~\citep{oord2019representationlearningcontrastivepredictive} variational bounds~\citep{poole2019variationalboundsmutualinformation} between latent actions and ground-truth action targets.
The examples are split into 2,560 training, 512 validation, and 1,024 test samples.
All mutual-information estimates are reported on the held-out test split.

For the BA estimator, we train an MLP conditional decoder $q_\phi(A\mid Z)$.
The decoder predicts the mean of a diagonal Gaussian over the action target, with a learned per-dimension log-standard deviation.
The BA score is computed as the held-out conditional log-likelihood improvement over a diagonal Gaussian marginal baseline fitted on the training action targets:
\[
    \widehat{I}_{\mathrm{BA}}(Z;A)
    =
    \mathbb{E}[\log q_\phi(A\mid Z)]
    -
    \mathbb{E}[\log p_0(A)].
\]
The implementation computes this quantity with natural logarithms; for reporting, we convert BA from nats to bits by dividing by $\ln 2$.   

For the InfoNCE estimator, we train separate MLP encoders for $Z$ and $A$.
The encoders map each input to a normalized 32-D embedding, and the similarity between sample $i$ and target $j$ is the scaled dot product between the two embeddings.
We use a symmetric contrastive objective, treating the matched pair $(Z_i,A_i)$ as positive and other samples in the batch as negatives:
\[
    \widehat{I}_{\mathrm{NCE}}
    =
    \log B
    -
    \frac{1}{2}
    \left[
    \mathrm{CE}(S,\mathrm{diag})
    +
    \mathrm{CE}(S^\top,\mathrm{diag})
    \right],
\]
where $B$ is the batch size and $S$ is the batch similarity matrix.
We report the held-out InfoNCE bound in natural-log units.

For probe diagnostics, we train supervised predictors from latent actions to the same action targets and report test $R^2$.
The probe uses same split as above.
For tokenized latent actions, we use an attentive pooling probe over latent tokens before the prediction head, so representations with different token counts can be evaluated under the same diagnostic.
The attentive probe applies train-split standardization, maps each token through a shared two-layer MLP with hidden dimension 128, computes a learned scalar attention score for each token, and predicts actions from the attention-weighted pooled representation.

\subsection{Wrong-Token Prediction Error}
\label{app:wrong_token_error}

We analyze token prediction errors using SimplerEnv-WidowX samples.
For each method, we load the VLA checkpoint after downstream fine-tuning and its corresponding fine-tuned action decoder checkpoint.
The frozen tokenizer provides the target latent action tokens, and the fine-tuned VLA predicts a distribution over valid action-token IDs.
We take the top-1 predicted token for each slot and compare it to the target token.

For the normalized wrong-token latent error reported in the main text, we map predicted and target tokens back to their latent code vectors and evaluate only token slots where the top-1 prediction is incorrect.
For each incorrect prediction, we compute the L2 distance between the predicted and target latent code vectors and normalize it by the average norm of the two code vectors.
We then average this normalized distance over all incorrect token predictions.
This measures whether a token mistake maps to a nearby latent code rather than an unrelated code.

To measure decoded action error, we use the actual fine-tuned VLA and action decoder rather than a separately trained proxy decoder.
We run the VLA with the target token sequence and again after replacing the latent action token positions with the VLA top-1 predictions.
The final-layer hidden states from these two passes are fed to the corresponding fine-tuned action decoder, and we measure the mean per-step L2 distance between the resulting normalized 10-step action chunks.

\subsection{Multi-Horizon Latent Policy}
\label{app:multi_horizon}

For the SimplerEnv-WidowX VLA experiment, the multi-horizon target concatenates latent action tokens for horizons $h\in\{3,6,9\}$ from the same observation window.
Each horizon contributes one radial token and four direction tokens, resulting in 15 latent action tokens in total.
For the continuous diffusion-policy experiments on Coffee and Stack Three, the multi-horizon target concatenates continuous latent actions for horizons $h\in\{5,10,20\}$.
All compared methods use the same downstream demonstrations and optimization budget within each task; only the latent target construction changes.

For Coffee and Stack Three, we train the continuous latent action pretraining stage and the latent policy stage for 15k steps each.
We then use joint diffusion-policy fine-tuning for 5k steps on Coffee and 3k steps on Stack Three.
For the SimplerEnv-WidowX VLA experiment, we evaluate the multi-horizon variant using matched 5k-step downstream fine-tuning checkpoints.

For the gradient-cosine diagnostic, we use SimplerEnv-WidowX samples and compute horizon-specific gradients from the per-horizon branch action L1 losses.
Table~\ref{tab:app_cross_horizon_grad} reports pairwise cosine values for all horizon pairs and their mean.

\begin{table}[t]
\centering
\caption{\textbf{Cross-horizon gradient cosine similarity.}
We report pairwise cosine similarity between gradients from losses for horizons $h\in\{3,6,9\}$ on SimplerEnv-WidowX.}
\label{tab:app_cross_horizon_grad}
\begin{tabular}{lcccc}
\toprule
\textbf{Method} & $\boldsymbol{h=3,6}$ & $\boldsymbol{h=3,9}$ & $\boldsymbol{h=6,9}$ & \textbf{Mean} \\
\midrule
PoLAR & \textbf{0.490} & \textbf{0.342} & \textbf{0.626} & \textbf{0.486} \\
\emph{Flat} ($5\times16$)& 0.409 & 0.133 & 0.373 & 0.305 \\
\bottomrule
\end{tabular}
\end{table}

\subsection{Hyperbolic versus Euclidean Diagnostics}

To compare hyperbolic and Euclidean PoLAR, we use matched tokenizers trained with the same data, architecture, radial objectives, and downstream setup, changing only the geometry used for radial losses.
We evaluate two diagnostics.

First, we measure adjacent-horizon direction similarity by averaging the cosine similarity between latent codes from the same start observation at adjacent future offsets.
Second, we remove radial scale from the latent action embedding and train the same attentive action probe used in the action-informativeness diagnostics at horizon $h=9$.

\section{Dataset and Evaluation Details}
\label{app:dataset_evaluation}

Table~\ref{tab:app_dataset_counts} summarizes the number of demonstrations used for pretraining, downstream fine-tuning, and analysis.

\begin{table}[t]
\centering
\caption{\textbf{Dataset counts.}}
\label{tab:app_dataset_counts}
\begin{tabular}{llc}
\toprule
\textbf{Dataset / task} & \textbf{Source} & \textbf{\# Demonstrations} \\
\midrule
Can & RoboMimic PH & 200 \\
Square & RoboMimic PH & 200 \\
Stack & MimicGen D0 & 1,000 \\
Mug Cleanup & MimicGen D0 & 1,000 \\
Threading & MimicGen D0 & 1,000 \\
Coffee & MimicGen D0 & 1,000 \\
Stack Three & MimicGen D0 & 1,000 \\
BridgeData V2 & BridgeData V2 & 60,096 \\
SimplerEnv-WidowX & our collection & 50 per task \\
Pick \& Place Banana & our collection & 50 \\
Cup Stack & our collection & 100 \\
Open Pot \& Banana & our collection & 100 \\
\bottomrule
\end{tabular}
\end{table}

\subsection{RoboMimic and MimicGen}

We use Can and Square from RoboMimic PH~\citep{robomimic2021}, and Stack, Mug Cleanup, Threading, Coffee, and Stack Three from MimicGen D0~\citep{mandlekar2023mimicgen}.
These tasks cover pick-and-place, stacking, object cleanup, insertion, and threading-style manipulation behaviors.
The main diffusion-policy experiments evaluate Can, Square, Stack, Mug Cleanup, and Threading; Coffee and Stack Three are used for the multi-horizon analysis.
All tasks use RGB observations and low-level 7-DoF robot actions.
RoboMimic \& MimicGen use 7-DoF Cartesian end-effector delta actions: 3D translational deltas, axis-angle rotational deltas, and a gripper command.
For evaluation, the diffusion policy predicts 20-step action sequences and executes the first 10 actions before replanning.
We report success over 100 rollout episodes per task.

\subsection{BridgeData V2}

We use BridgeData V2~\citep{walke2023bridgedata} as the large-scale pretraining dataset for the discrete latent action and VLA experiments.
The dataset provides top-view RGB observations, language instructions, and robot actions from WidowX manipulation trajectories.
BridgeData V2 uses the standard Bridge action representation consisting of a world-frame translation delta, a rotation delta, and a gripper command.
BridgeData V2 is used only for pretraining; downstream evaluation is conducted on SimplerEnv-WidowX and real-world robot tasks.

\subsection{SimplerEnv-WidowX}

For SimplerEnv-WidowX~\citep{li24simpler}, we evaluate four downstream tasks: Put Spoon on Towel, Put Carrot on Plate, Stack Green Block on Yellow Block, and Put Eggplant in Basket.
For each task, we fine-tune on 50 successful demonstrations collected for that task and evaluate on 240 held-out episodes that do not overlap with the fine-tuning demonstrations.
Simplerenv-WidowX also uses 7-DoF Cartesian end-effector delta actions.
At evaluation time, each policy predicts a 10-step action chunk at every environment step.
We keep an overlapping buffer of recent predicted chunks: before adding a new chunk, existing chunks are advanced by one timestep, and the new chunk is inserted as the most recent prediction.
The executed action is a normalized exponential average over all valid current-step predictions in the buffer, with weights $\exp(-0.1 i)$ for chunk age $i$.
Thus the newest chunk receives the largest weight, while older overlapping predictions still contribute when valid.
All VLA methods on SimplerEnv-WidowX, including all baselines, use the same 10-step action-chunk horizon and temporal aggregation procedure.

\subsection{Real-World Robot Tasks}

We evaluate three real-world tasks on WidowX SoloAI robot platforms: Pick \& Place Banana, Cup Stack, and Open Pot \& Banana.
In Pick \& Place Banana, the robot picks up a banana and places it at the target location.
In Cup Stack, the robot picks up either cup and places it into the other cup.
In Open Pot \& Banana, the robot first opens the pot lid and then picks and places the banana into the pot.
We collect demonstrations through WidowX leader-follower teleoperation using the LeRobot framework~\citep{cadene2026lerobotopensourcelibraryendtoend}.
Demonstration collection and policy evaluation both use fixed top-view RGB observations captured by an Intel RealSense Depth Camera D435.
Unlike the simulation and SimplerEnv-WidowX experiments, which use Cartesian end-effector delta actions, our real-world LeRobot demonstrations use 7-DoF absolute joint-position action targets from the leader-follower teleoperation interface.
Each method is fine-tuned on demonstrations collected on the same robot platform.
At evaluation time, policy inference runs on an NVIDIA DGX Spark, and the policy predicts 10-step action chunks and executes all 10 actions open-loop before replanning.
We run 30 trials per task.
For sequential tasks, we report success after each required stage as well as final task success, and compute the overall average using final task success rates.



\section{Additional Results}
\subsection{Detailed Simulation Results}
Tables~\ref{tab:app_decoder_joint_finetune} and~\ref{tab:app_simplerenv_success} report the per-task success rates.
For RoboMimic and MimicGen, each success rate is averaged over 100 rollout episodes per task; for SimplerEnv-WidowX, each success rate is computed over 240 held-out evaluation episodes per task.

\begin{table*}[t]
    \centering
    \begin{minipage}[t]{0.48\textwidth}
        \centering
   \caption{\textbf{RoboMimic and MimicGen task success rates.}
Success rates (\%) are averaged over 100 episodes per task for both decoder-only and joint fine-tuning settings.}
        \label{tab:app_decoder_joint_finetune}
        \resizebox{\linewidth}{!}{%
        \begin{tabular}{lcccccc}
            \toprule
            \textbf{Method} & \textbf{Can} & \textbf{Square} & \textbf{Stack} & \textbf{Mug} & \textbf{Threading} & \textbf{Avg.} \\
            \midrule\midrule
            \multicolumn{7}{l}{\textbf{Decoder Only}} \\
            \midrule
            \rowcolor{blue!20}
            PoLAR (Hyp) & \textbf{92.0} & \textbf{61.0} & \textbf{80.0} & \textbf{22.0} & \textbf{21.0} & \textbf{55.2} \\
            PoLAR (Euc) & 85.0 & 53.0 & 48.0 & 12.0 & \textbf{21.0} & 43.8 \\
            \emph{Flat} & 37.0 & 49.0 & 73.0 & 3.0  & 9.0  & 34.2 \\
            \midrule\midrule
            \multicolumn{7}{l}{\textbf{Joint Fine-tuning}} \\
            \midrule
            \rowcolor{blue!20}
            PoLAR (Hyp) & 74.0 & 71.0 & \textbf{87.0} & \textbf{56.0} & \textbf{63.0} & \textbf{70.2} \\
            PoLAR (Euc) & 29.0 & 57.0 & 57.0 & 37.0 & 1.0  & 36.2 \\
            \emph{Flat ($5\times16$)} & 44.0 & 49.0 & 69.0 & 11.0 & 12.0 & 37.0 \\
            ResNet18    & \textbf{76.0} & \textbf{81.0} & 79.0 & 3.0 & 30.0 & 53.8 \\
            \bottomrule
        \end{tabular}
        }
    \end{minipage}
    \hfill
    \begin{minipage}[t]{0.48\textwidth}
        \centering
        \caption{\textbf{SimplerEnv-WidowX task success rates.}
        We report success rates (\%) across four manipulation tasks and their average.}
        \label{tab:app_simplerenv_success}
        \resizebox{\linewidth}{!}{%
        \begin{tabular}{lccccc}
            \toprule
            \textbf{Method} & \textbf{Spoon} & \textbf{Carrot} & \textbf{Stack} & \textbf{Eggplant} & \textbf{Avg.} \\
            \midrule\midrule
            \multicolumn{6}{l}{\textbf{PoLAR and variants}} \\
            \midrule
            \rowcolor{blue!20}
            PoLAR                         & 72.9 & \textbf{62.9} & 36.3 & \textbf{77.1} & \textbf{62.3} \\
            PoLAR (\emph{dir-only})       & 70.8 & 44.2 & 31.3 & 41.3 & 46.9 \\
            \emph{Fact.}                  & 64.6 & 49.2 & 31.3 & 49.6 & 48.6 \\
            \emph{Flat} ($4\times32$) & 67.9 & 45.4 & 24.2 & 40.0 & 44.4 \\
            \emph{Flat} ($5\times16$) & 71.3 & 45.4 & 30.8 & 52.1 & 49.9 \\
            \midrule\midrule
            \multicolumn{6}{l}{\textbf{Latent action baselines}} \\
            \midrule
            UniVLA~\citep{bu2025univla}      & 68.8 & 42.9 & 32.9 & 47.9 & 48.1 \\
            Villa-X~\citep{chen2025villa0x0} & 61.3 & 43.8 & 6.7 & 36.7 & 37.1 \\
            LAPA~\citep{ye2024latent}        & 60.0 & 50.8 & 20.8 & 50.0 & 45.4 \\
            \midrule\midrule
            \multicolumn{6}{l}{\textbf{Pretrained VLA references}} \\
            \midrule
            $\pi_{0.5}$~\citep{intelligence2025pi05visionlanguageactionmodelopenworld} & \textbf{73.8} & 55.4 & 34.6 & 62.5 & 56.6 \\
            SmolVLA~\citep{shukor2025smolvlavisionlanguageactionmodelaffordable}       & 56.7 & 40.8 & \textbf{37.9} & 53.8 & 47.3 \\
            \bottomrule
        \end{tabular}
        }
    \end{minipage}
\end{table*}

\subsection{Real-World Rollouts}

\begin{figure*}[t]
    \centering
    \includegraphics[width=0.9\textwidth]{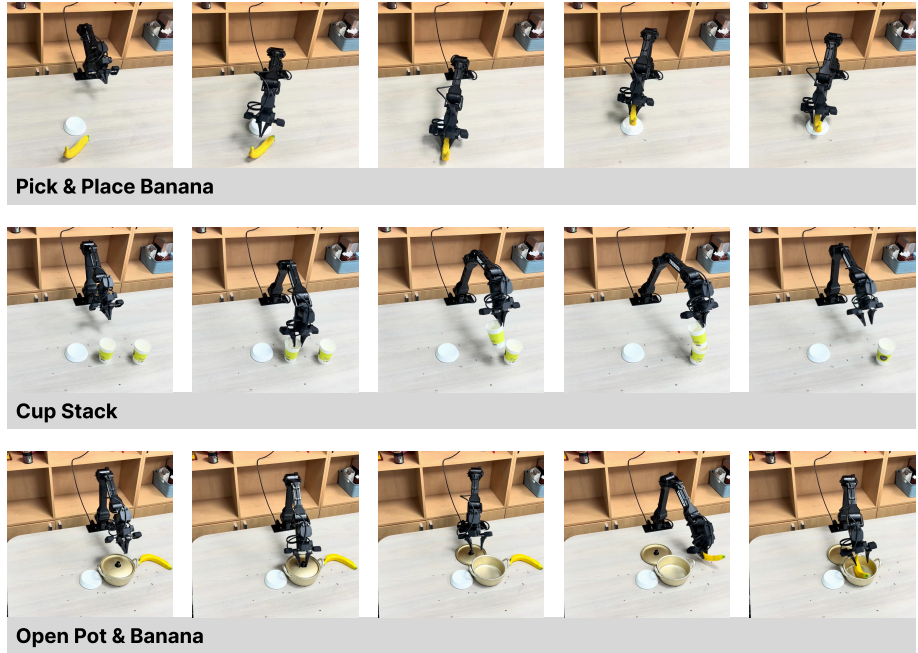}
    \caption{\textbf{Successful real-world PoLAR rollouts.}
    We show representative successful executions for Pick \& Place Banana, Cup Stack, and Open Pot \& Banana.
    Each row shows temporally ordered snapshots from one rollout.}
    \vspace{-1.5em} 
    \label{fig:app_real_world_success}
\end{figure*}

Figure~\ref{fig:app_real_world_success} shows representative successful PoLAR rollouts on the three real-world tasks.
The snapshots illustrate the full task progression for Pick \& Place Banana, Cup Stack, and Open Pot \& Banana under the same camera and robot setup used for evaluation.

\begin{figure*}[t]
    \centering
    \includegraphics[width=0.9\textwidth,keepaspectratio]{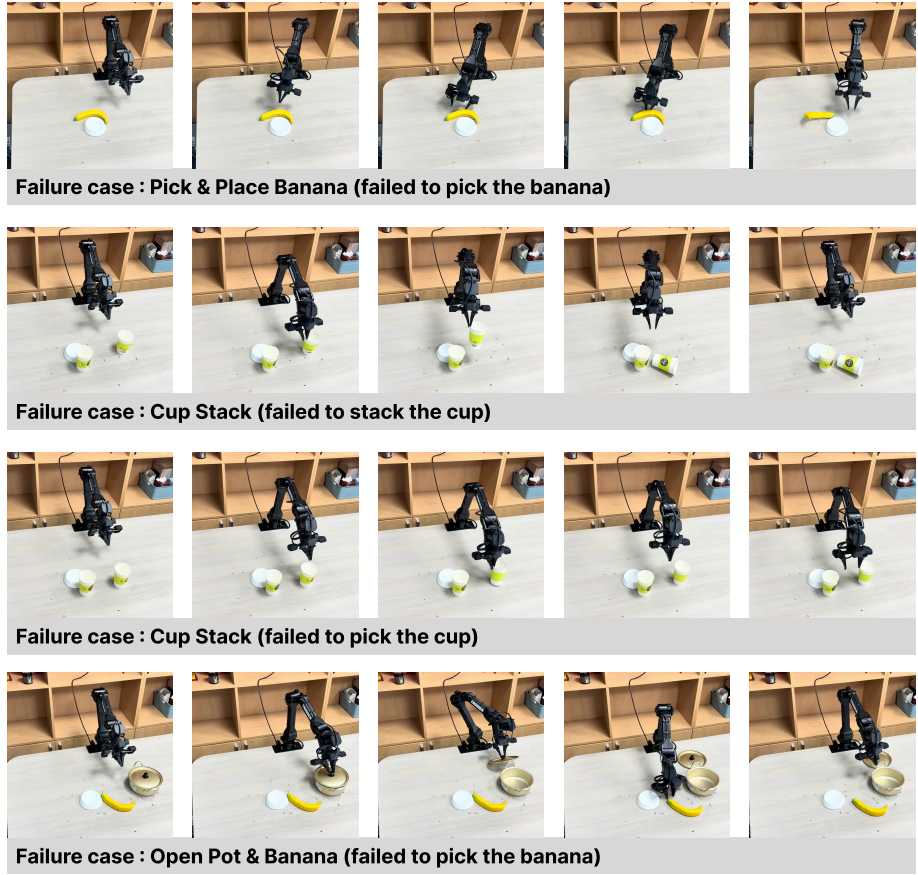}
    \caption{\textbf{Observed PoLAR failure cases in real-world rollouts.}
    We include all observed PoLAR failures from the real-world evaluation.
    The failures mainly arise from grasping errors, unsuccessful cup stacking, or failure to complete the banana pick stage in the sequential pot task.}
    \label{fig:app_failure_polar}
    \vspace{-1.5em}
\end{figure*}

\begin{figure*}[t]
    \centering
    \includegraphics[width=0.9\textwidth,keepaspectratio]{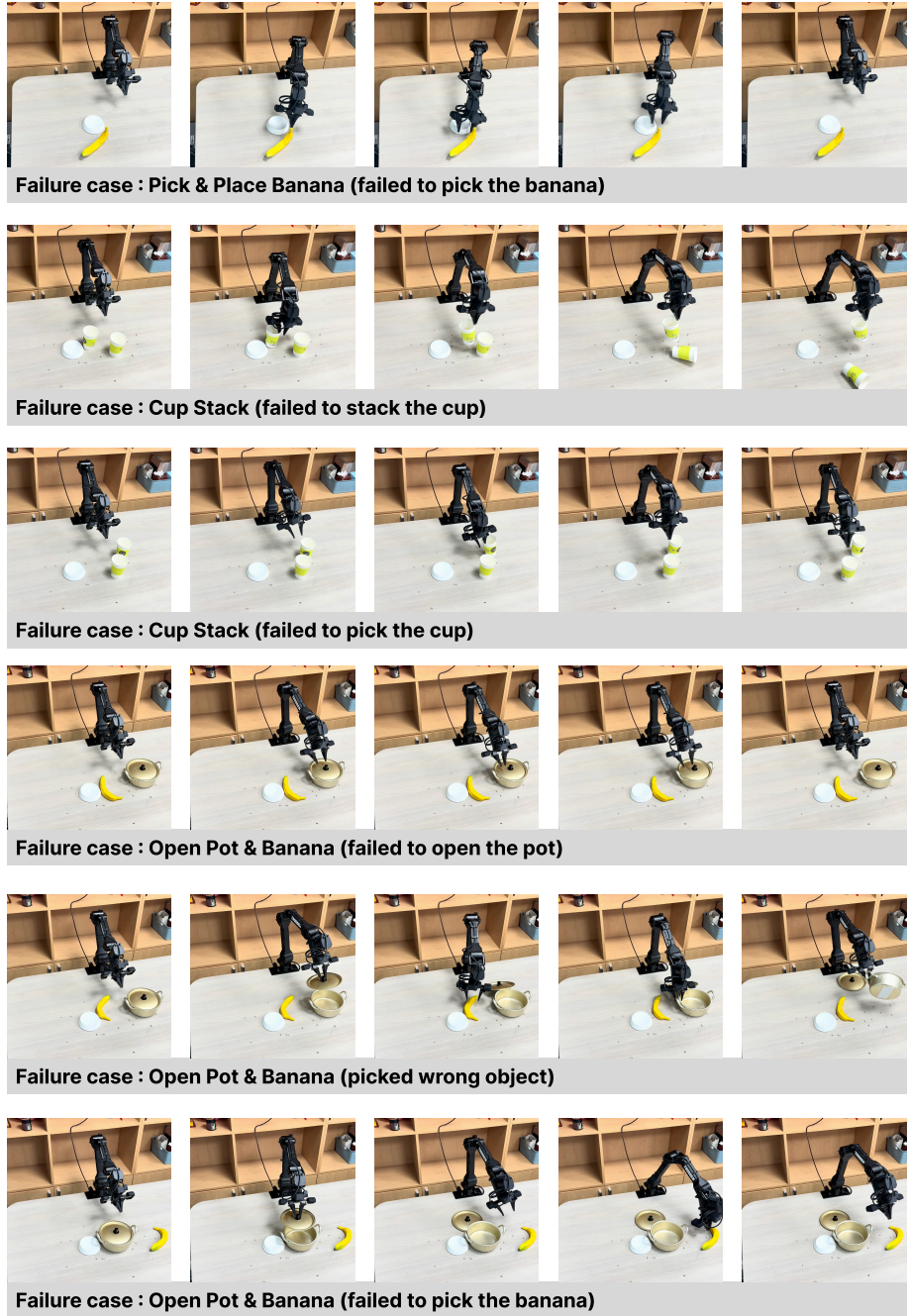}
    \caption{\textbf{Observed baseline failure cases in real-world rollouts.}
    We include all observed baseline failures from the real-world evaluation.
    Compared with PoLAR, the baseline failures cover a broader set of errors, including grasp failures, unsuccessful cup stacking, failure to open the pot, selecting the wrong object, and failure to complete the banana pick stage.}
    \label{fig:app_failure_baselines}
    \vspace{-1.5em}
\end{figure*}

\paragraph{Failure cases.}
Figures~\ref{fig:app_failure_polar} and~\ref{fig:app_failure_baselines} show the observed real-world failure cases.
For PoLAR, failures include unsuccessful banana grasps,  failing to stack or pick a cup, and failing to pick and place the banana after opening the pot.
For the baseline methods, the observed failures additionally include failing to open the pot and selecting the wrong object.
These figures summarize the observed failure modes from the corresponding real-world rollouts.

\subsection{Additional Radius Sweep Visualizations}
\label{app:radius_sweep_more}

\begin{figure*}[t]
    \centering
    \includegraphics[width=0.95\textwidth]{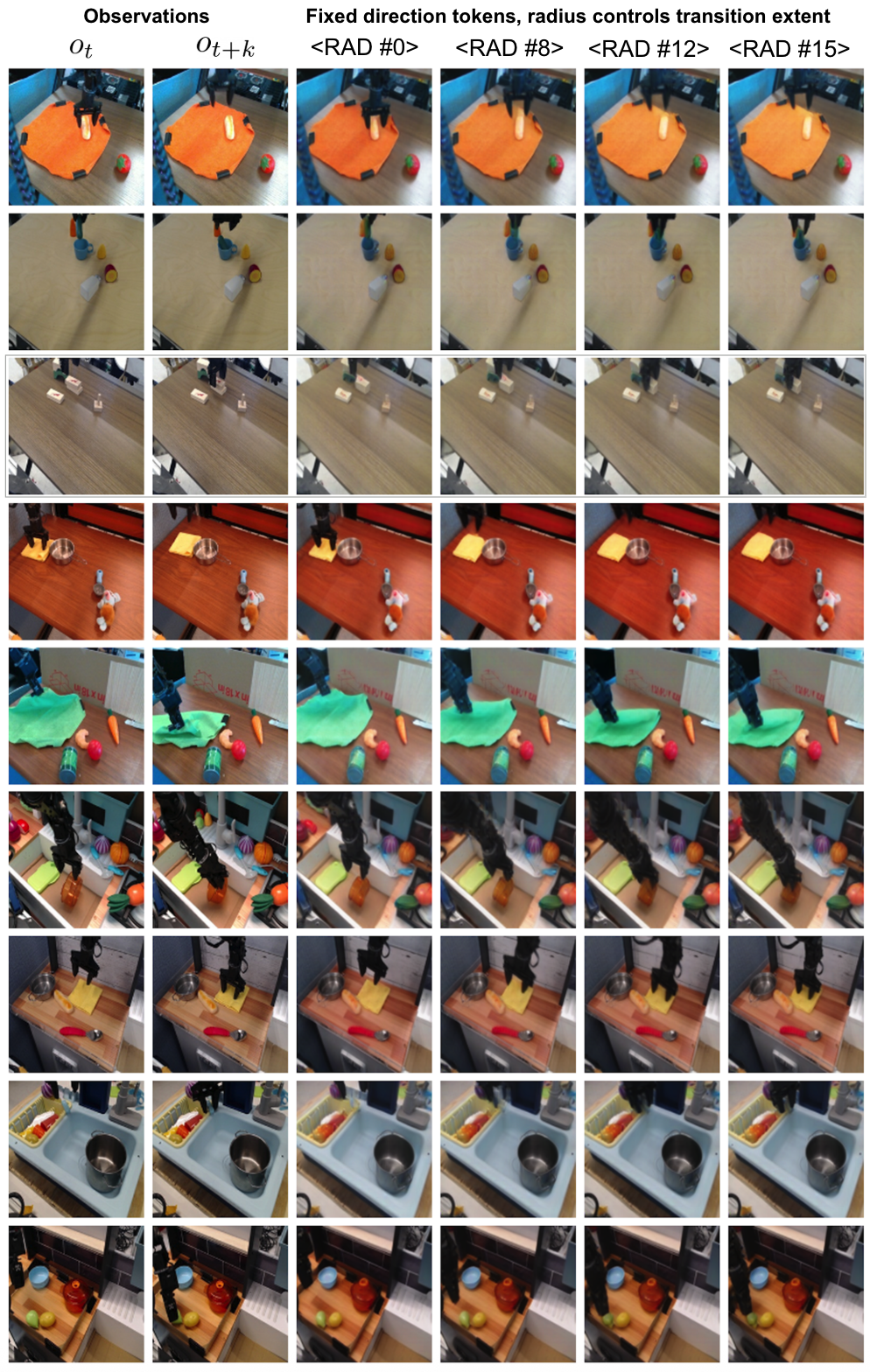}
    \caption{\textbf{Additional radius sweep visualizations.}
    We fix the direction tokens and vary only the radial token.
    Larger radii generally produce larger decoded visual transitions while preserving the transition mode.}
    \label{fig:app_radius_sweep_more}
\end{figure*}

Figure~\ref{fig:app_radius_sweep_more} shows additional radius-sweep examples.
We fix the direction tokens and vary only the radial token before decoding the FDM-predicted DINOv2 features.
Across examples, increasing radius generally produces larger visual transitions while preserving the transition mode.

\end{document}